\documentclass[conference]{IEEEtran}

\usepackage[pdftex]{graphicx}

\usepackage[cmex10]{amsmath}
\usepackage{amssymb}

\usepackage[caption=false,font=footnotesize]{subfig}

\usepackage{url}
\usepackage{xcolor}
\usepackage{supertabular}
\usepackage{arydshln}

\usepackage{tikz}

\begin{document}

\title{Learning scale-variant and scale-invariant features for deep image 
classification}

\author{\IEEEauthorblockN{Nanne van Noord and Eric Postma}
\IEEEauthorblockA{Tilburg center for Communication and Cognition, School of Humanities \\
Tilburg University, The Netherlands\\
Email: \{n.j.e.vannoord, e.o.postma\}@tilburguniversity.edu}}

\maketitle

\begin{abstract}
Convolutional Neural Networks (CNNs) require large image corpora to be trained 
on classification tasks.  The variation in image resolutions, sizes of objects 
and patterns depicted, and image scales, hampers CNN training and performance, 
because the task-relevant information varies over spatial scales.
Previous work attempting to deal with such
scale variations focused on encouraging scale-invariant CNN representations.
However, scale-invariant representations are incomplete representations of 
images, because images contain scale-variant information as well.
This paper addresses the combined development of scale-invariant and
scale-variant representations. We propose a multi-scale CNN method to
encourage the recognition of both types of features and evaluate it on a
challenging image classification task involving task-relevant characteristics
at multiple scales. The results show that our multi-scale CNN outperforms
single-scale CNN. This leads to the conclusion that encouraging the combined
development of a scale-invariant and scale-variant representation in CNNs is
beneficial to image recognition performance. \end{abstract}

\section{Introduction} \label{sec:intro}
Convolutional Neural Networks (CNN) have drastically
changed the computer vision landscape by considerably improving the performance
on most image benchmarks \cite{Krizhevsky2012, He2015}. A key characteristic of 
CNNs is that the deep(-based) representation, used to perform the classification, 
is generated from the data, rather than being engineered. The deep 
representation determines the type of visual features that are used for 
classification. In the initial layers of the CNN, the visual features 
correspond to oriented edges or color transitions. In higher layers, the visual 
features are typically more complex (e.g., conjunctions of edges or shapes). 
Finding the appropriate representation for the task at hand requires presenting the 
CNN with many instances of a visual entity (object or pattern) in all its 
natural variations, so that the deep representation captures most naturally 
occurring  appearances of the entity.  

Three main sources of natural 
variation are the location, the viewpoint, and the size of an object or 
pattern.  Variations in location are dealt with very well by a 
CNN~\cite{Gong2014}, which follows naturally from the weight sharing employed 
in the convolution layers \cite{LeCun1995a}. CNNs can also handle variations in 
viewpoint by creating filters that respond to viewpoint-invariant features 
\cite{Kheradpisheh2015}.  Size variations pose a particular challenge in CNNs 
\cite{Xu2014}, especially when dealing with image corpora containing images of 
varying resolutions and depicting objects and patterns at different sizes and 
scales, as a result of varying distances from the camera and blurring by 
optical imperfections, respectively. This leads to variations in image 
resolution, object size, and image scale, which are two different properties of 
images.  The relations between image resolution, object size, and image scale 
is formalized in digital image analysis using Fourier theory 
\cite{Gonzalez:2006:DIP:1076432}. Spatial frequencies are a central concept in 
the Fourier approach to image processing. Spatial frequencies are the 
two-dimensional analog of frequencies in signal processing. The fine details of 
an image are captured by high spatial frequencies, whereas the coarse visual 
structures are captured by low spatial frequencies. In what follows, we provide 
a brief intuitive discussion of the relation between resolution and scale, 
without resorting to mathematical formulations.

\subsection{Image resolution, object size, and image scale}
Given an image its resolution can be expressed in terms of the number of pixels 
(i.e., the number of samples taken from the visual source); low resolution 
images have fewer pixels than high resolution images. The scale of an image 
refers to its spatial frequency content. Fine scale images contain the range 
from high spatial frequencies (associated with small visual structures) down to 
low spatial frequencies (with large visual structures). Coarse scale images 
contain low spatial frequencies only. The operation of spatial smoothing (or 
blurring) of an image corresponds to the operation of a low-pass filter: high 
spatial frequencies are removed and low spatial frequencies are retained. So, 
spatial smoothing a fine scale image yields a coarser scale image.

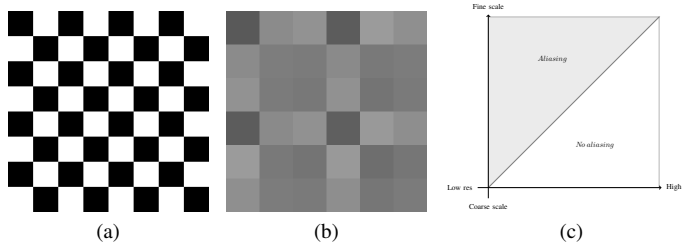
\begin{figure}[!t]
  \centering
  \subfloat[]{
    \includegraphics[width=.3\linewidth]{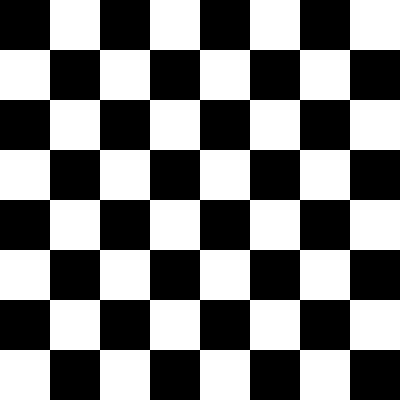}
  }
  \subfloat[]{
    \includegraphics[width=.3\linewidth]{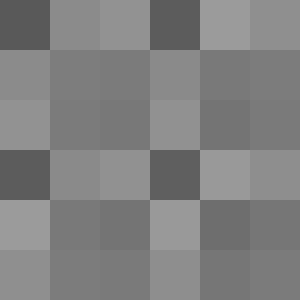}
  }
  \subfloat[]{ \resizebox{.38\linewidth}{!}{\begin{tikzpicture}[
        dot/.style = {
	      draw,
	      fill = white,
	      circle,
	      inner sep = 0pt,
	      minimum size = 4pt
	    }
	  ]
	
    \node[label=below:Coarse scale] (C) at (0,-0.5) {};
    \node[label=above:Fine scale] (F) at (0,8.1) {};

    \node[label=left:Low res] (LR) at (-0.5,0) {};
    \node[label=right:High res] (HR) at (8.1,0) {};
	
    \draw[->,thick] (LR.center) -- (HR.center);
    \draw[->,thick] (C.center) -- (F.center);
	
    \draw[thick,black,fill=gray!50!white, opacity=0.3] (0,0) -- (8,8) -- (0,8) 
    -- (0,0);
    \node[] (AS) at (3,6) {$Aliasing$};
    \draw[thick,black,opacity=0.3] (0,0) -- (8,8) -- (8,0) -- (0,0);
    \node[] (NAS) at (5,2) {$N\! o\: aliasing$};
	
  \end{tikzpicture} }  }
  \caption{Illustration of aliasing. (a) Image of a chessboard. (b) Reproductions of the chessboard with an image of insufficient resolution ($6 \times 6$ pixels). The reproduction is obtained by applying bicubic interpolation. (c) The space spanned by image resolution and image scale. Images defined by resolution-scale combinations in the shaded area suffer from aliasing. See text for details.}\label{fig:chessboard}
\end{figure}

The relation between the resolution and the scale of an image follows from the 
observation that in order to represent visual details, an image should have a 
resolution that is sufficiently high to accommodate the representation of the 
details. For instance, we consider the chessboard pattern shown in 
Figure~\ref{fig:chessboard}a. Figure~\ref{fig:chessboard}b shows a $6 \times 6$ 
pixel reproduction of the chessboard pattern. The resolution of the 
reproduction is insufficient to represent the fine structure of the chessboard 
pattern. The distortion of an original image due to insufficient resolution (or 
sampling) is called {\em aliasing} \cite{Gonzalez:2006:DIP:1076432}.

As this example illustrates, image resolution imposes a limit to the scale 
at which visual structure can be represented.  Figure~\ref{fig:chessboard}c 
displays the space spanned by resolution (horizontal axis) and scale (vertical 
axis). The limit is represented by separation of the shaded and unshaded 
regions.  Any image combining a scale and resolution in the shaded area suffers 
from aliasing. The sharpest images are located at the shaded-unshaded boundary.  
Blurring an image corresponds to a vertical downward movement into the unshaded 
region

\begin{figure*}[!t]
  \centering
    \includegraphics[width=0.6\linewidth]{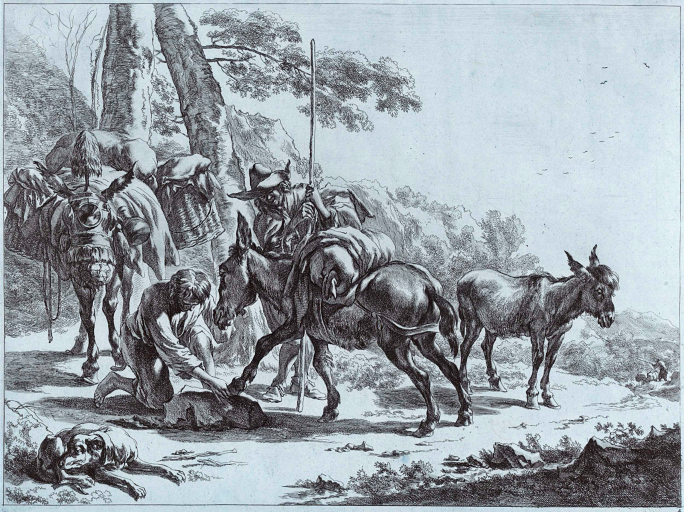}
  \caption{Artwork \textit{`Hoefsmid bij een ezel'} by Jan de 
Visscher.}\label{fig:artwork}
\end{figure*}

Having discussed the relation between resolution and scale, we now turn to the 
discussion of the relation of object size to resolution and scale. Real-world 
images with a given scale and resolution contain objects and structures at a 
range of sizes \cite{Lindeberg1994}, For example, the image of the artwork 
shown in Figure~\ref{fig:artwork}, depicts large-sized objects (people and 
animals) and small-sized objects (hairs and branches). In addition, it may 
contain visual texture associated with the paper it was printed on and with the 
tools that were used used to create the artwork. Importantly, the same object 
may appear at different sizes. For instance, in the artwork shown there persons 
depicted at different sizes. The three persons in the middle are much larger in 
size than the one at the lower right corner. The relation between image 
resolution and object size is that the resolution puts a lower bound on the 
size of objects that can be represented in the image. If the resolution is too 
low, the smaller objects cannot be distinguished anymore. Similarly, the 
relation between image scale and object size is that if the scale becomes too 
coarse, the smaller objects cannot be distinguished anymore. Image smoothing 
removes the high-spatial frequencies associated with the visual characteristics 
of small objects. 

\subsection{Scale-variant and scale-invariant image representations}

Training CNNs on large image collections that often exhibit variations in image 
resolution, depicted object sizes, and image scale, is a challenge. The 
convolutional filters, which are automatically tuned during the CNN training 
procedure, have to deal with these variations.  Supported by the acquired 
filters the CNN should ignore task-irrelevant variations in image resolution, 
object size, and image scale and take into account task-relevant features at a 
specific scale. The filters providing such support are referred to as 
scale-invariant and scale-variant filters, respectively \cite{Gluckman2006}.

The importance of scale-variance was previously highlighted by Gluckman 
\cite{Gluckman2006} and Park et al. \cite{Park2010}, albeit for two different 
reasons. The first reason put forward by Gluckman arises from the observation 
that images are only partially described by scale invariance 
\cite{Gluckman2006}. When decomposing an image into its scale-invariant 
components, by means of a scale-invariant pyramid, and subsequently 
reconstructing the image based on the scale-invariant components the result 
does not fully match the initial image, and the statistics of the resulting 
image do not match those of natural images. For training a CNN this means that 
when forcing the filters to be scale-invariant we might miss image structure 
which is relevant to the task. Gluckman demonstrated this, by means of his 
proposed space-variant image pyramids, which separate scale-specific from 
scale-invariant information in \cite{Gluckman2006} and found that object 
recognition benefited from scale-variant information.  

The second reason was presented by Park et al. in \cite{Park2010}, where they 
argue that the need for scale-variance emerges from the limit imposed by image 
resolution, stating that ``Recognizing a 3-pixel tall object is fundamentally 
harder than recognizing a 300-pixel object or a 3000-pixel object.'' \cite[p.  
2]{Park2010}. While recognising very large objects comes with it own 
challenges, it is obvious that the recognition task can be very different 
depending on the resolution of the image. Moreover, the observation that 
recognition changes based on the resolution ties in with the previously 
observed interaction between resolution and scale: as a reduction in resolution 
also changes the scale. Park et al. identify that most multi-scale models 
ignore that most naturally occurring variation in scale, within images, occurs 
jointly with variation in resolution, i.e. objects further away from the camera 
are represented at a lower scale and at a lower resolution.  As such they 
implement a multi-resolution model and demonstrate that explicitly 
incorporating scale-variance boosts performance.

Inspired by these earlier studies, we propose a multi-scale CNN which 
explicitly deals with variation in resolution, object size and image scale, by 
encouraging the development of filters which are scale-variant, whilst 
constructing a representation that is scale-invariant.

The remainder of this paper is organised as follows.  Section~\ref{sec:relwork} 
contains an overview of previous work that deals with scale variation for 
learning deep image representations.  In Section~\ref{sec:method} we provide a 
detailed presentation of our multi-scale CNN for scale-invariant and 
scale-variant filters.  Section~\ref{sec:task} outlines the task used for 
evaluating the performance of the multi-scale CNN.
In Section~\ref{sec:exp} the experimental setup is described, including the 
dataset and the experimental method.
In Section~\ref{sec:results} the results of the experiments are presented. We 
discuss the implications of using multi-scale CNNs in 
Section~\ref{sec:discussion}.  Finally, Section~\ref{sec:conclusion} concludes 
by stating that combining scale-variant and scale-invariant features 
contributes to image classification performance.

\section{Previous work}
\label{sec:relwork}
In this paper, we examine learning deep image representations that incorporate 
scale-variant and/or scale-invariant visual features by means of CNNs.  Scale 
variation in images and its impact on computer vision algorithms is a widely 
studied problem \cite{Lindeberg1994, Lowe2004}, where invariance is often 
regarded as a key property of a representation \cite{Lenc2015}. It has been 
shown that under certain conditions CNN will develop scale-invariant filters 
\cite{Le2010}. Additionally, various authors have investigated explicitly 
incorporating scale-invariance in deep representations learnt by CNN 
\cite{Sermanet2013, Xu2014, Gong2014, Kanazawa, jaderberg2015}. While these 
approaches successfully deal with scale-invariance they forgo the problem of 
recognising scale-variant features at multiple scales \cite{Park2010}.

\textit{Standard} CNN trained without any data augmentation will develop 
representations which are scale-variant. As such it is only capable of 
recognising the features it was trained on, at the scale it was trained on, 
such a CNN cannot deal with scale-variant features at different scales. A 
straightforward solution to this limitation is to expose the CNN to multiple 
scales during training, this approach is typically referred to as scale 
jittering \cite{Szegedy2014, Simonyan2015,Girshick2015}. It is commonly used as 
a data augmentation approach to increase the amount of training dataset, and as 
a consequence reduce overfitting. Additionally, it has been shown that scale 
jittering improves classification performance \cite{Simonyan2015}. While part 
of the improved performance is due to the increase in training data and reduced 
overfitting, scale jittering also allows the CNN to learn to recognise more 
scale-variant features, and potentially develop scale-invariant filters.  
Scale-invariant filters might emerge from the CNN being exposed to scale 
variants of the same feature. However, \textit{standard} CNN typically do not 
develop scale-invariant filters \cite{Le2010}, and instead will require more 
filters to deal with the scaled variants of the same feature \cite{Xu2014}, in 
addition to the filters needed to capture scale-variant features.
A consequence of this increase in parameters, which increases further when more 
scale variation is introduced, is that the CNN becomes more prone to overfit 
and training the network becomes more difficult in general. In practice this 
limits scale-jittering to small scale variations. Moreover, scale-jittering is 
typically implemented as jittering the resolution, rather than explicitly 
changing the scale, which potentially means that jittered versions are actually 
of the same scale.

One approach that is able to deal with larger scale variations, whilst offering 
many of the same benefits as scale jittering is multi-scale training 
\cite{Wu2015}. Multi-scale training consists of training separate CNN on fixed 
size crops of resized versions of the same image. At test time the softmax 
class posteriors of these CNN are averaged into a single prediction, taking 
advantage of the information from different scales and model averaging 
\cite{Ciresan2012a}, resulting in improved performance over single scale 
classification.  However, because the work by Wu et al. \cite{Wu2015} is 
applied to datasets with a limited image resolution, they only explore the 
setting in which multi-scale training is applied for a relatively small 
variation in scales, and only two scales.  Moreover, as dealing with scale 
variation is not an explicit aim of their work they do not analyse the impact 
of dealing with multiple scales, beyond that it increases their performance.
Finally, because of the limited range of scales they explored they do not deal 
with aliasing due to resizing. Aliasing is harmful for any multi-scale approach 
as it produces visual artifacts which would not occur in natural images of the 
reduced scale, whilst potentially obfuscating relevant visual structure at that 
scale.

In this work we aim to explicitly learn scale-variant features for large 
variations in scale, and make the following three contributions: (1) We present 
a modified version of multi-scale training that explicitly creates multiple 
scales, reducing aliasing due to resizing, allowing us to compare larger scale 
differences whilst reducing redundancy between scales.  (2) We introduce a 
novel dataset of high resolution images that allows us to explore the effects 
of larger scale variations.  (3) We perform an in-depth analysis of the results 
and compare different scale combinations in order to increase our understanding 
of the influence of scale-variation on the classification performance.

\section{Multi-scale Convolutional Neural Network}
\label{sec:method}
In this Section we present the multi-scale CNN by explaining how a standard 
(single-scale) CNN performs a spatial decomposition of images. Subsequently, we 
motivate the architecture of the multi-scale CNN in terms of the 
scale-dependency of the decomposition. 

CNNs perform a stage-wise spatial decomposition of the input, for an image of a 
face this is typically described in terms of pixels which combine into edges, 
which combine into contours, into simple parts of faces, and finally into 
entire faces. This is achieved by repeating alternating convolution and pooling 
operations across stages. At the first stage, in the convolution operation, the 
image is transformed by a set of several (learned) filters with a limited 
spatial extent (typically a small sub-region of the image). After which the 
pooling operation reduces the dimensionality of the convolution.  At each 
subsequent convolution-pooling stage, the output of the previous stage is 
convolved by another set of (learned) filters and subsequently pooled 
\cite{LeCun1995a}. As a consequence, both the complexity of the composite 
transformation and the image area covered increases with each stage 
\cite{Garcia2004}. Therefore, relatively simple visual patterns with a small 
spatial extent are processed at the early stages, whereas more complex visual 
patterns with a large spatial extent are processed at the later stages 
\cite{LeCun1995a, Xu2014}. This dependency closely ties the representation and 
recognition of a visual pattern to its spatial extent, and thus to a specific 
stage in the network \cite{Sermanet2011, Hariharan2014}. 

\begin{figure}[!t]
\centering
\includegraphics[width=2.5in]{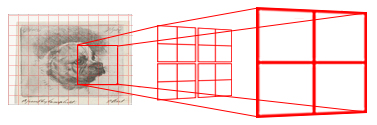}
\caption{A $2 \times 2$ filter applied to the output of $4$ filters of the 
same size in a lower layer corresponds to a $4 \times 4$ region in the input 
image.}
\label{fig:convscale}
\end{figure}

The strength of this dependency is determined by the network architecture in 
which the amount of subsampling (e.g., via strided operations or pooling) is 
specified, this also determines the size of the spatial output of the network.
In the case of a simple two layer network with $2 \times 2$ filters as in 
Figure~\ref{fig:convscale}, the network produces a single spatial output per $4 
\times 4$ region in the input. Whereas in a deeper network (containing strided 
and pooling operations such as in \cite{Krizhevsky2012}) a single output can 
describe a $64 \times 64$ pixel region of the input.  Because the amount of 
subsampling is determined by the network architecture, the size of the output, or 
spatial output map, scales with the size of the input. Due to the scaling the 
relative portion of the input described by a single output node decreases: a $4 
\times 4$ pixels image can be described with $4$ non-overlapping $2 \times 2$ 
filters, where each filter describes one-fourth of the image.  Yet for an $8 
\times 8$ image it would require $16$ identically sized filters to cover the 
input, reducing the portion of the image described by each filter to 
one-sixteenth.  The reduction in relative proportion described by a single 
output strongly influences the characteristics of the filters in the network.  
Filters that describe one-sixteenth of a portrait picture might only correspond 
to a part of a nose, or an ear, whereas filters that cover one-fourth of the 
picture might correspond to an entire cheek, chin, or forehead.  For artist 
attribution this means that a network with filters that cover relatively small 
parts of the input are suitable to describe the fine characteristics but cannot 
describe the composition or iconography of the artwork. As such the network 
architecture should be chosen in concurrence with the resolution of the input.

Because training CNNs on an image dataset results in a hierarchy of feature 
representations with increasing  spatial extent, a network capable of analysing 
the entire range from fine to coarse visual characteristics in an image 
requires many stages in order to capture all the intermediate scales.  
Moreover, as to not discard information by subsampling between stages, the 
subsampling has to be performed gradually.  Gradual subsampling is performed by 
having a very deep network with many stages, each subsampling a little.  The 
complexity and the number of parameters in a network is determined by the 
number of layers and the number of parameters per layer, as such, increasing 
the number of layers increases the complexity of the network. A more complex 
network requires more training data, which despite the increasing availability 
of images of artworks is still lacking.  Moreover, the computational demand of 
the network increases strongly
with the complexity of the network, making it infeasible to train a 
sufficiently complex network \cite{Hou2015}. An alternative to increasing the 
complexity of an individual CNN is to distribute the task over specialised CNNs 
and combining the resulting predictions into a single one.  The biologically 
motivated multi-column CNN architecture \cite{Ciresan2012} is an example of 
such an approach. 

\begin{figure}[!t]
\centering
\includegraphics[width=\linewidth]{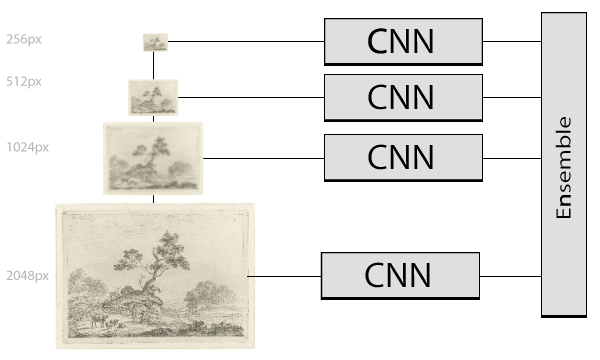}
\caption{Visual representation of the model architecture.}
\label{fig:model}
\end{figure}

The multi-scale CNN presented in this paper is based on a multi-scale image representation, 
whereby a separate CNN is associated with each scale. This allows the scale-specific CNNs to 
develop both scale-variant and scale-invariant features. The multi-scale 
representation is created using a Gaussian pyramid \cite{Adelson1984}.  The 
bottom level of the pyramid corresponds to the input image, subsequent levels 
contain smoothed (and down-sampled) versions of the previous levels. A visual 
representation of the model architecture is shown in Figure~\ref{fig:model}.
Note that down-sampling is not necessary to create the higher pyramid levels, 
and that it is possible to fix the resolution and only change the scale.  
However, smoothing results in a redundancy between neighbouring pixels, as they 
convey the same information. 

\section{Image classification task}
\label{sec:task}
The proposed multi-scale CNN will be evaluated on a task involving a large data 
set of images of artworks that are heterogeneous in scale and resolution. In 
our previous work, we have applied a single CNN to a comparable dataset to 
study computational artist attribution (where the task was to determine who 
authored a given artwork) \cite{Noord2015}.  For artist attribution there is 
often insufficient information on a single scale to distinguish between very 
similar artists.  For instance, the works of two different artists who use very 
similar materials to create artworks depicting different scenes might be 
indistinguishable when considering the very fine details only.  Alternatively, 
when artists create artworks depicting a similar scene using different 
materials, these may be indistinguishable at a coarse spatial scale.
Hence, successful artist attribution requires scale-variant features in 
addition to scale-invariant features.

Artist attribution is typically performed by combining current knowledge on the 
artist's practices, technical data, and a visual assessment of the artwork as 
to establish its origin and value from an economical and historical perspective 
\cite{Johnson2008}.  In recent years it has been shown that this visual 
assessment can be performed computationally and can lead to promising results 
on artist attribution image classification tasks \cite{Taylor2007, Johnson2008, 
Hughes2010a, VanderMaaten2010, Li2012, Abry2013}.  The increased availability 
of visual data from the vast digital libraries of museums and the challenges 
associated with the unique nature of artworks has led to an interest in this 
domain by researchers from a large diversity of fields. This diversity has 
resulted in a great many different approaches and techniques aimed at tackling 
the problem of visual artist attribution.  The visual assessment of artworks by 
art experts generally focuses on textural characteristics of the surface (e.g., 
the canvas) or on the application method (e.g., brushstrokes) 
\cite{Hendriks2008}, this in turn has shaped many of the computational 
approaches to visual artwork assessment (e.g., \cite{Johnson2008, Hendriks2012, 
Li2012}). 

More recently it has been shown that general purpose computer vision approaches 
can be used for the visual assessment of artworks, specifically SIFT features 
\cite{Mensink2014} and deep-based representations as learned by a CNN for a 
general object recognition task (i.e., ImageNet) \cite{Karayev2013, Saleh2015} 
can be used to perform image classification tasks on artworks.
This development is a deviation from the practice as performed by art experts, 
with the focus shifted from small datasets of a few artists with high 
resolution images ($5$ to $10$ pixels per mm) to large datasets with many 
artists and lower resolution images ($0.5$ to $2$ pixels per mm).  By using 
images of a lower resolution the amount of details related to the artist's 
specific style in terms of application method (e.g., brushstrokes) and material 
choices (e.g., type of canvas or paper) become less apparent, which shifts the 
focus to coarser image structures and shapes.  However, using a multi-scale 
approach to artist attribution it is possible to use information from different 
scales, learning features appropriate from both coarse and fine details.

\section{Experimental setup}
\label{sec:exp}
This section describes the setup of the artist attribution experiment. The 
setup consists of a specification of the CNN architecture, the dataset, the 
evaluation, and the training parameters.

\subsection{multi-scale CNN architecture}
The multi-scale CNN architecture used in this work is essentially an ensemble 
of single-scale CNN, where the single-scale CNN matches the architecture of the 
previously proven ImageNet model described in \cite{springenberg2015}.  We made 
two minor modifications to the architecture described in 
\cite{springenberg2015} in that we (1) replaced the final $6 \times 6$ average 
pooling layer with a global average pooling layer which averages the final 
feature map regardless of its spatial size, and (2) reduce the number of ouputs 
of the softmax layer to $210$ to match the number of classes in our dataset. A 
detailed specification of the single-scale CNN architecture can be found in 
Table~\ref{tab:netarch}, where conv-$n$ denotes a convolutional layer with $f$ 
filters with a size ranging from $11 \times 11$ to $1 \times 1$.  The stride 
indicates the step size of the convolution in pixels, and the padding indicates 
how much zero padding is performed before the convolution is applied. 

The single-scale CNN architecture used is fully-convolutional, which means that 
except for the final global average pooling layer it consists solely of 
convolutional layers. Rather than having max or average pooling layers in the 
network a convolutional layer with a stride greater than $1$ (typically $2$) is 
used.  This convolutional layer effectively performs the pooling, but combines 
it with an additional (learnt) non-linear transformation.
A fully convolutional architecture has two main benefits for the work described 
in this paper: (1) unlike traditional CNN, a fully-convolutional CNN places no 
restrictions on the input in terms of resolution; the same architecture can be 
used for varying resolutions, and (2) it can be trained on patches and 
evaluated on whole images, which makes training more efficient and evaluation 
more accurate. 

Additionally, this architecture has been shown to work well with Guided 
Backpropagation (GB) \cite{springenberg2015}. GB is an approach (akin to 
`deconvolution' \cite{Zeiler2014}) that makes it possible to visualise what the 
network has learnt, or which parts of an input image are most characteristic of 
a certain artist. GB consists of performing a backward pass through the network 
and computing the gradient w.r.t. an input image. In order to visualise which 
parts of an image are characteristic of an artist, the activations of the 
softmax class posterior layer are all set to zero, except the activation for 
the artist of interest, and subsequently the gradient w.r.t. an input image 
will activate strongest in the areas characteristic of that artist.

\begin{table}[!t]
\caption{CNN architecture of single-scale networks as used in this paper.  
  conv$n$ denote convolutional
  layers. During training a $224 \times 224$ pixels crop is used, the testing 
is performed on the entire input image (which shortest side is in the range of 
$256$ up to $2048$ pixels).}
\label{tab:netarch}
\centering
\begin{tabular}{c|c|c|c}
Layer & Filters & Size, stride, pad & Description \\
\hline
Training Data & - & $224 \times 224$, -, - & RGB image crop \\
\hdashline
Testing Data & - & Entire image, -, - & Full RGB image \\
\hline
conv1.1 & $96$ & $11 \times 11$, 4, 0 & ReLU \\ 
conv1.2 & $96$ & $1 \times 1$, 1, 0 & ReLU \\
conv1.3 & $96$ & $3 \times 3$, 2, 1 & ReLU \\ 
\hline 
conv2.1 & $256$ & $5 \times 5$, 1, 2 & ReLU \\
conv2.2 & $256$ & $1 \times 1$, 1, 0 & ReLU \\ 
conv2.3 & $256$ & $3 \times 3$, 2, 0 & ReLU \\
\hline
conv3.1 & $384$ & $3 \times 3$, 1, 1 & ReLU \\
conv3.2 & $384$ & $1 \times 1$, 1, 0 & ReLU \\
conv3.3 & $384$ & $3 \times 3$, 2, 0 & ReLU + Dropout ($50\%$) \\
\hline
conv4 & $1024$ & $1 \times 1$, 1, 0 & ReLU \\
conv5 & $1024$ & $1 \times 1$, 1, 0 & ReLU \\
conv6 & $210$ & $1 \times 1$, 1, 0 & ReLU \\
\hline
global-pool & - & - & Global average \\
softmax & - & - & Softmax layer \\
\end{tabular}
\end{table}

Our multi-scale is constructed as an ensemble, or multi-column 
\cite{Ciresan2012a}, architecture, in which the softmax class-posteriors of the 
single-scale CNN are averaged and used as the final predictions for evaluation, 
the evaluation procedure is further described in Section~\ref{subsec:eval}.

\subsection{Dataset}
\begin{figure}[!t]
\centering
\includegraphics[width=2.5in]{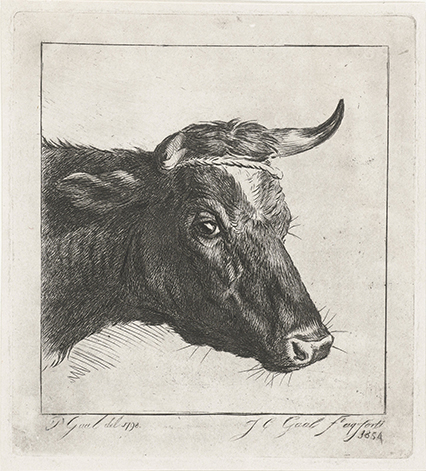}
\caption{Digital photographic reproduction of 
  \textit{`Kop van een koe met touw om de horens'} by Jacobus Cornelis Gaal.}
\label{fig:mopshond}
\end{figure}

The dataset\footnote{The dataset is available at \url{https://auburn.uvt.nl/}.} 
consists of $58,630$ digital photographic reproductions of print artworks by 
$210$ artists retrieved from the collection of the Rijksmuseum, the Netherlands 
State Museum.  These artworks were chosen based on the following four criteria: 
(1) Only printworks made on paper, (2) by a single artist, (3) public domain, 
and (4) at least $96$ images by the same artist match these criteria.
This ensured that there were sufficient images available from each artist to 
learn to recognise their work, and excluded any artworks which are visually 
distinctive due to the material choices (e.g., porcelain).  An example of a 
print from the Rijksmuseum collection is shown in Figure~\ref{fig:mopshond}.  

\begin{figure}[!t]
\centering
\includegraphics[width=2.8in]{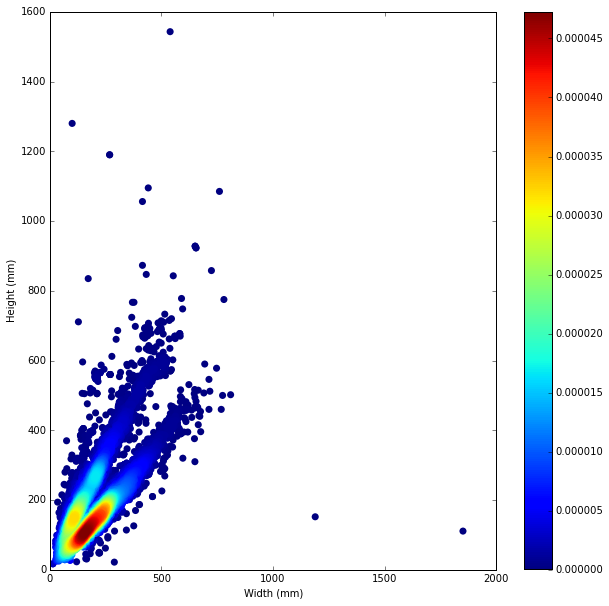}
\caption{Scatter plot of physical dimensions of the artworks in the test set in 
millimeters; each point represents an artwork, its colour indicating the 
density in the area around it. The scatter plot shows that there are two 
predominant shapes of artworks: square artworks and rectangular artworks  
(width slightly greater than height). The majority of the artworks cluster 
around a size of $250 \times 250$ mm. }
\label{fig:sizedist}
\end{figure}

For many types of artworks there is a large degree of variation in their 
physical size: there are paintings of several meters in width or height, and 
paintings which are only tens of centimeters
in width or height. Moreover, for such artworks there is a large degree of 
variation in the ratio of pixels per mm and as such the dimension of the 
reproductions in pixels. Yet, this makes
it very appealing to work with print artworks, as they are much more uniform in 
terms of physical size as for example paintings. While there is still some 
variation in physical size for print artworks, as shown in 
Figure~\ref{fig:sizedist}.
Previous approaches have dealt with such variations by resizing all images to a 
single size, which confounds image resolution with physical resolution. 

Normalising the images to obtain fixed pixel to mm ratios would result in a 
loss of visual detail. Given that our aim is to have our multi-scale CNN 
develop both scale-invariant and scale-variant filters, we take the variation 
in scales and resolutions for granted.

A four-level Gaussian (low-pass) pyramid is created following the standard 
procedure for creating Gaussian Pyramids described in \cite{Adelson1984, 
Ogden1985}.  Initially all images are resized so that the shortest side (height 
or width) is $2048$ pixels, as to preserve the aspect ratio, creating the first 
pyramid level. From this first level the subsequent pyramid level is created by 
smoothing the previous level, and down-sampling by removing every other pixel 
column and row (effectively reducing the image size by a factor two).  This 
smoothing and down-sampling step is repeated, every time taking the previous 
level as the starting point, to create the remaining two pyramid levels.
The smoothing steps were performed by recursively convolving the images with 
the Gaussian kernel $G$, which is defined as:
\begin{displaymath}
G = \frac{1}{256} \begin{bmatrix} 1 & 4 & 6 & 4 & 1  \\ 4 & 16 & 24 & 16 & 4  \\ 6 
& 24 & 36 & 24 & 6  \\ 4 & 16 & 24 & 16 & 4  \\ 1 & 4 & 6 & 4 & 1 \end{bmatrix}
\end{displaymath}. 

The resulting Gaussian pyramid consists of four levels of images with 
the shortest side being $256$, $512$, $1024$, and $2048$ pixels for each level 
respectively.

The dataset is divided into a training ($70\%$), validation ($10\%$), and test 
set ($20\%$). The training set is used to
train the network, the validation set is used to optimise the hyperparameters, 
and the evaluation set is used to estimate the prediction performance.  All 
results reported in this paper are based on the test set.


\subsection{Training parameters}
All networks were trained using an effective training procedure (cf. 
\cite{Krizhevsky2012}), with the values of the learning rate, momentum, and 
weight decay hyperparameters being $10^{-2}$, $0.9$, and $5 \cdot 10^{-4}$ 
respectively. Whenever the error on the validation set stopped decreasing the 
learning rate was decreased by a factor $10$. 

\subsection{Evaluation}
\label{subsec:eval}
The evaluation is performed on entire images. The fully-convolutional nature of 
the multi-scale CNN makes it unnecessary to perform cropping. The 
scale-specific prediction for an image is the average over the spatial output 
map, resulting in a single scale-specific prediction for the entire image. The 
performance on all experiments is reported using the Mean Class Accuracy (MCA), 
which is the average of the accuracy scores obtained per artist. We report the 
MCA because it is not sensitive to unbalanced classes and it allows for a 
comparison of the results with those reported in \cite{Mensink2014, Noord2015}.  
The MCA is equal to the mean of the per class precision, as such we also report 
the mean of the per class recall, and the harmonic mean of these mean precision 
and mean recall measures, also known as the F-score.

Additionally, we compare our results to those obtained by performing 
multi-scale training as described in \cite{Wu2015}.
We implemented multi-scale training using the same CNN architecture as used 
previously, and only varied the input data. Rather than blurring the images 
before subsampling the images, we follow \cite{Wu2015} and directly subsample 
the images, as such the scales do not form a Gaussian Pyramid.  Because the 
highest scale is not blurred in either case these results are identical, and 
are produced by the same network.  

Furthermore, we report the pair-wise correlations between the Class Accuracy 
(CA) for each artist for the four different scales for both approaches. The 
pair-wise correlations between scales indicates the similarity of the 
performance for individual artists at those two scales. A high correlation 
indicates that the attributions of an artist are largely the same at both 
scales, whereas a low correlation indicates that the artworks of an artists are 
classified differently at the two scales which suggests the relevance of 
scale-specific information.

\section{Results}
\label{sec:results}
The results of each individual scale-specific CNN of the multi-scale CNN and 
the ensemble averages are reported in Table~\ref{tab:results}.  The 
best-performing single scale is $512$. The ensemble-averaged score of the 
multi-scale CNN outperforms each individual scale by far. As is evident from 
Table~\ref{tab:subresults}, no combination of three or fewer scales outperforms 
the multi-scale (four-scale) CNN. We report the results obtained by multi-scale 
training \cite{Wu2015} in Table~\ref{tab:wu_results}. The results of all 
possible combinations of these results are reported in 
Appendix~\ref{app:subresults}.

\begin{table}[!t]
\caption{Mean Class Accuracies, Mean recall and F-score for the four individual 
scales and the ensemble of four scales for our approach.}
\label{tab:results}
\centering
\begin{tabular}{cccc}
\hline
  Scale & MCA  & Mean recall & F-Score \\
\hline
$256$ & $70.36$ & $65.03$ & $67.59$ \\
$512$ & $75.69$ &  $69.70$ & $72.57$ \\
$1024$ & $67.69$ & $44.08$ & $53.48$ \\
$2048$ & $62.03$ & $38.54$ & $47.55$ \\
Ensemble & $\mathbf{82.11}$ &  $\mathbf{72.50}$ &  $\mathbf{77.01}$ \\
\hline
\end{tabular}
\end{table}

\begin{table}[!t]
  \caption{Mean Class Accuracies for all possible scale combinations obtained 
  with our approach, a `+' indicates inclusion of the scale. In bold are the 
combinations which lead to the best combined performance in each block. The 
best overall score is underlined.}
\label{tab:subresults}
\centering
\begin{tabular}{cccc|ccc}
\hline
256 & 512 & 1024 & 2048 & MCA & Mean recall & F-Score\\
\hline
+ & {} & {} & {} & $70.36$ & $65.03$ & $67.59$ \\
{} & + & {} & {} & $\mathbf{75.69}$ & $\mathbf{69.7}$ & $\mathbf{72.57}$ \\
{} & {} & + & {} & $67.96$ & $44.08$ & $53.48$ \\
{} & {} & {} & + & $62.03$ & $38.54$ & $47.55$ \\
\hline
+ & + & {} & {} & $78.06$ & $\mathbf{71.61}$ & $\mathbf{74.69}$ \\
+ & {} & + & {} & $75.92$ & $67.65$ & $71.54$ \\
+ & {} & {} & + & $76.24$ & $67.92$ & $71.84$ \\
{} & + & + & {} & $79.15$ & $67.71$ & $72.98$ \\
{} & + & {} & + & $\mathbf{80.21}$ & $68.11$ & $73.66$ \\
{} & {} & + & + & $71.41$ & $45.4$ & $55.51$ \\
\hline
+ & + & + & {} & $80.15$ & $72.14$ & $75.94$ \\
+ & + & {} & + & $80.87$ & $\mathbf{72.47}$ & $\mathbf{76.44}$ \\
+ & {} & + & + & $79.27$ & $68.89$ & $73.72$ \\
{} & + & + & + & $\mathbf{80.95}$ & $65.9$ & $72.66$ \\
\hline
+ & + & + & + & $\mathbf{\underline{82.12}}$ & $\mathbf{\underline{72.5}}$ & $\mathbf{\underline{77.01}}$ \\
\hline
\end{tabular}
\end{table}

\begin{table}[!t]
\caption{Mean Class Accuracies, Mean recall and F-score for the four individual 
scales and the ensemble of four scales using multi-scale training 
\cite{Wu2015}.}
\label{tab:wu_results}
\centering
\begin{tabular}{cccc}
\hline
  Scale & MCA  & Mean recall & F-Score \\
\hline
$256$ & $70.56$ & $65.74$ & $68.07$ \\
$512$ & $73.5$ & $68.36$ & $70.84$ \\
$1024$ & $65.63$ & $57.96$ & $61.56$ \\
$2048$ & $62.03$ & $38.54$ & $47.55$ \\
Ensemble & $\mathbf{79.98}$ &  $\mathbf{73.02}$ &  $\mathbf{76.34}$ \\
\hline
\end{tabular}
\end{table}

The MCA and mean recall obtained for the resolutions greater than $512$ 
decrease, this suggests that there is a ceiling in performance and that further 
increasing the resolution would not help to improve the performance.  Yet, 
combining the predictions from each scale in an ensemble results in a boost in 
performance. The pair-wise correlations between scales as reported in 
Table~\ref{tab:corresults} show larger correlations for adjacent scales than 
for non-adjacent scales. This pattern of correlations agrees with the causal 
connection of adjacent scales. Additionally, we also report the correlations 
between the scales using multi-scale training (c.f.  \cite{Wu2015}) in 
Table~\ref{tab:wu_corresults}. We note that in general the correlations in the 
latter case are stronger than the former, which shows that there is a greater 
performance difference across artists between scales for our approach, which 
indicates that the single-scale CNN for our approach learn a greater variety of 
scale-variant features.

\begin{table}[!t]
\caption{Correlations between results per artist for each image scale}
\label{tab:corresults}
\centering
\begin{tabular}{lrrrr}
\hline
{} &  256 &  512 &  1024 &  2048  \\
\hline
256  &        1.00 &            0.56 &            0.27  &       0.18        \\
512  &        0.56 &            1.00 &            0.44  &       0.29        \\
1024 &        0.27 &            0.44 &            1.00  &       0.54        \\
2048 &        0.18 &            0.29 &            0.54  &       1.00        \\
\hline
\end{tabular}
\end{table}

\begin{table}[!t]
\caption{Correlations between results per artist for each image scale using 
  multi-scale training \cite{Wu2015}.}
\label{tab:wu_corresults}
\centering
\begin{tabular}{lrrrr}
\hline
{} &  256 &  512 &  1024 &  2048  \\
\hline
256  &        1.00 &        0.60 &         0.33 &         0.26 \\
512  &        0.60 &        1.00 &         0.52 &         0.35 \\
1024 &        0.33 &        0.52 &         1.00 &         0.40 \\
2048 &        0.26 &        0.35 &         0.40 &         1.00 \\
\hline
\end{tabular}
\end{table}

To provide some insight on artist-specific relevance of the four different 
scales, Table~\ref{tab:topflop} lists the top five artists with the least and 
most variation between scales as determined by the standard deviation of their 
MCA across scales. From this
table it can be observed that there is a large variation between artists in 
terms of which scales work well, where for some artists performance is highly 
scale-specific (a perfect performance is achieved on one scale and a completely
flawed performance on another), and for others performance does not depend on 
scale (the performance is stable across scales).

\begin{table}[!t]
\caption{Overview of artists with the least and most variation between scales,
and their MCA per scale.}
\label{tab:topflop}
\centering
\begin{tabular}{lrrrr}
\hline
\multicolumn{5}{c}{Top five artists with least variation between scales.} \\
\hline
Artist &  256 &  512 &  1024 &  2048  \\
\hline
Johannes Janson & $66.67$ & $65.12$ & $67.74$ & $65.22$ \\
Pieter de Mare & $80.0$ & $82.67$ & $86.0$ & $81.25$ \\
Jacobus Ludovicus Cornet & $73.53$ & $76.47$ & $73.33$ & $79.17$ \\
Cornelis van Dalen (II) & $100.0$ & $94.44$ & $100.0$ & $100.0$ \\
Lucas Vorsterman (I) & $85.42$ & $89.8$ & $83.67$ & $88.57$ \\
\hline
\hline
\multicolumn{5}{c}{Top five artists with most variation between scales.} \\
\hline
Artist &  256 &  512 &  1024 &  2048  \\
\hline
Joannes van Doetechum (I) & $100.0$ & $100.0$ & $0.0$ & $0.0$ \\
Totoya Hokkei & $100.0$ & $0.0$ & $0.0$ & $0.0$ \\
Gerrit Groenewegen & $88.89$ & $100.0$ & $100.0$ & $0.0$ \\
Abraham Genoels & $86.67$ & $64.29$ & $0.0$ & $0.0$ \\
Charles Meryon & $64.0$ & $86.67$ & $100.0$ & $0.0$ \\
\hline
\end{tabular}
\end{table}

\begin{figure*}[!t]
  \centering
  \subfloat[Art work at $256$]{
    \includegraphics[width=.24\linewidth]{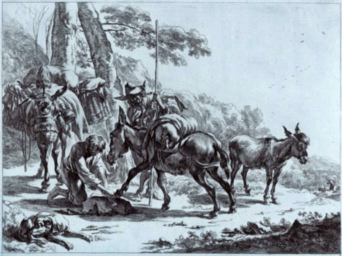}
  }
  \subfloat[Activation]{
    \includegraphics[width=.24\linewidth]{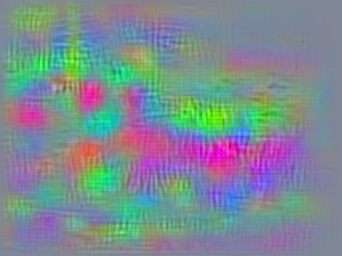}
  }
  \subfloat[Art work at $512$]{
    \includegraphics[width=.24\linewidth]{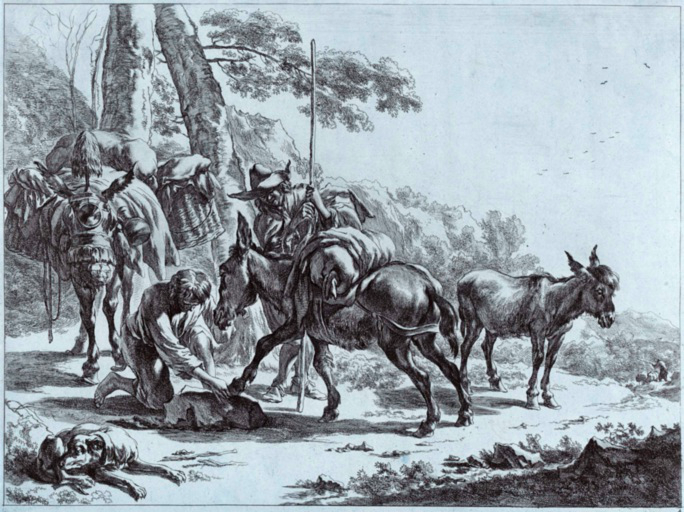}
  }
  \subfloat[Activation]{
    \includegraphics[width=.24\linewidth]{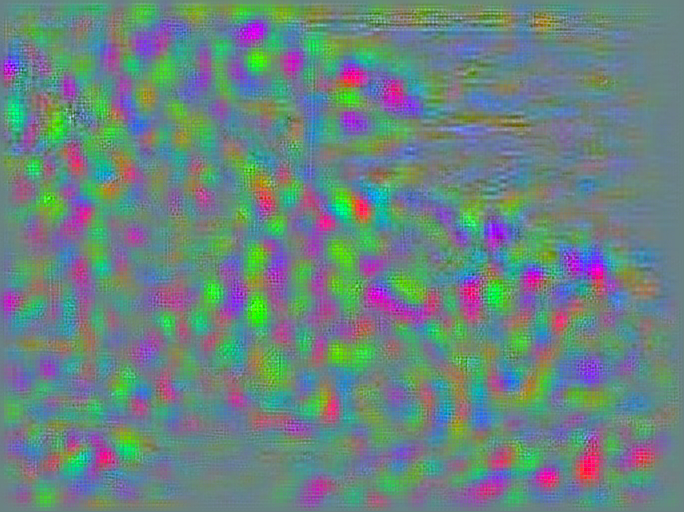}
  }
  \hfil
  \subfloat[Art work at $1024$]{
    \includegraphics[width=.24\linewidth]{1024_pyr_gbp8494_70original.png}
  }
  \subfloat[Activation]{
    \includegraphics[width=.24\linewidth]{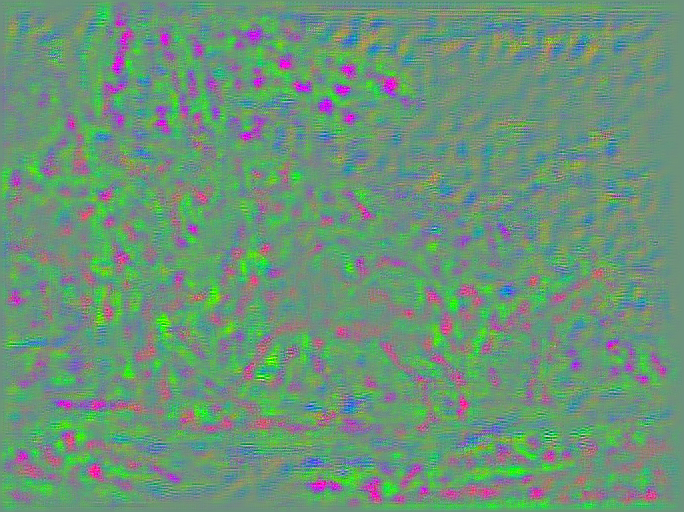}
  }
  \subfloat[Art work at $2048$]{
    \includegraphics[width=.24\linewidth]{1024_pyr_gbp8494_70original.png}
  }
  \subfloat[Activation]{
    \includegraphics[width=.24\linewidth]{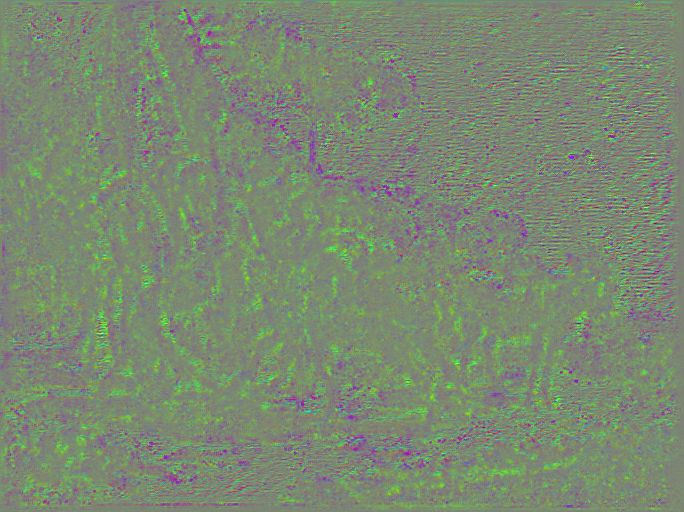}
  }
  \caption{Visualisations of the activations for the artwork \textit{`Hoefsmid 
    bij een ezel'} by Jan de Visscher at four scales. The activation shows the 
    importance of the highlighted regions for correctly identifying the artist, 
    the colours have been contrast enhanced for increased visibility. Best 
  viewed in colour.}\label{fig:guidedbackp}
\end{figure*}

To illustrate the effect of resolution on the automatic detection of 
artist-specific features, Guided Backpropagation \cite{springenberg2015} was 
used to create visualisations of the artwork \textit{`Hoefsmid bij een ezel'} 
by Jan de Visscher at the four scales. Figure~\ref{fig:guidedbackp} shows the 
results of applying Guided Backpropagation to the art work. The visualisations 
show the areas in the input image that the network considers characteristic of 
Jan de Visscher for that scale. A clear shift to finer details is observed when 
moving to higher resolutions.

\begin{figure*}[!t]
\centering
\includegraphics[width=\linewidth]{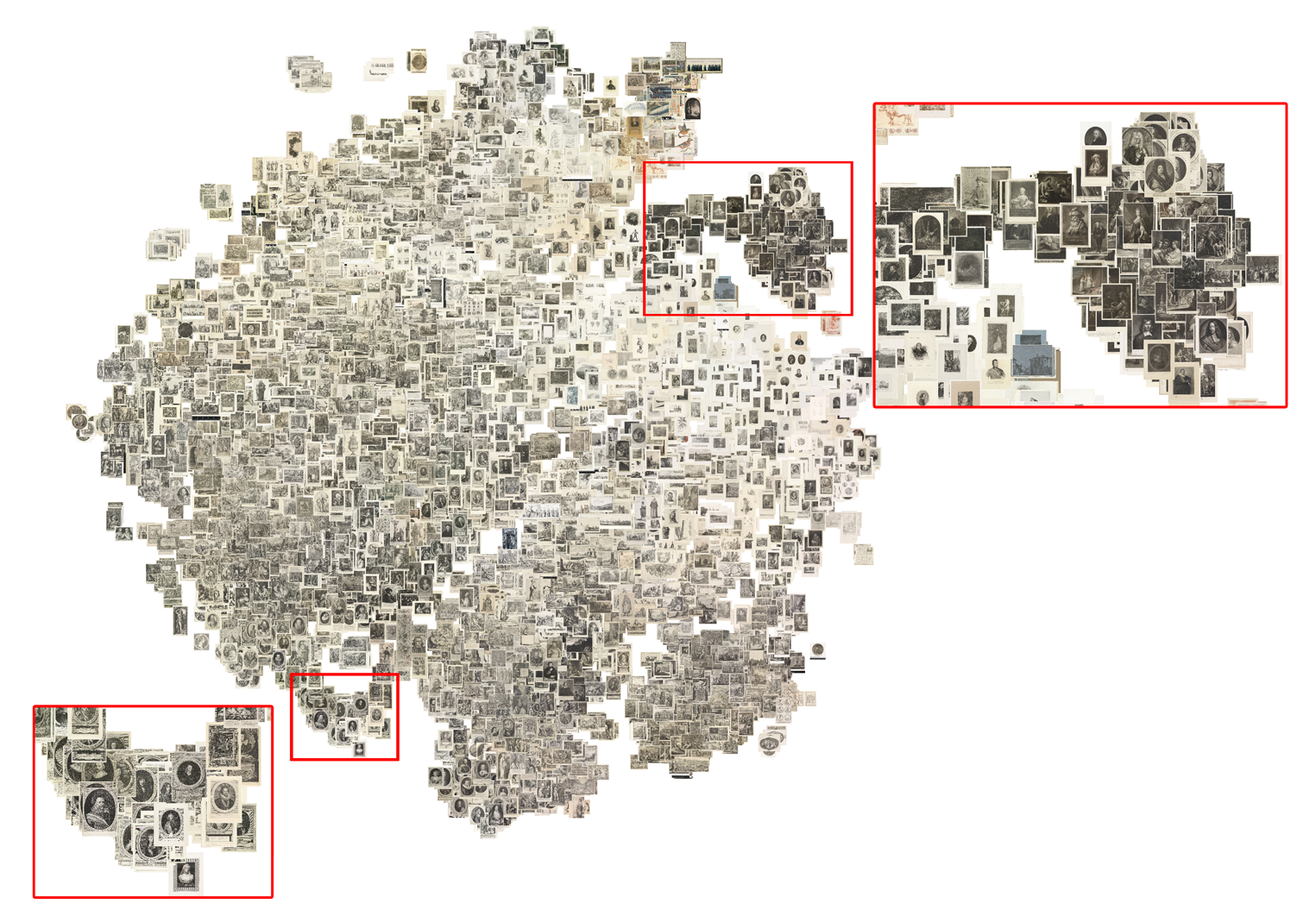}
\caption{t-SNE plot of all artworks in the test set where spatial distance 
indicates the similarity as observed by the network. Zoomed excerpts shown of 
outlined areas, illustrating examples of highly similar clusters.}
\label{fig:tsnefull}
\end{figure*}

As the multi-scale CNN produces a prediction vector for each image we are able 
to calculate the similarity of the artworks in terms of the distance in a 
high-dimensional space. Using t-SNE \cite{Maaten2008}
we visualise these distances in a two-dimensional space in 
Figure~\ref{fig:tsnefull}, the spatial distance indicates the similarity 
between images at determined by the ensemble.  The t-SNE visualisation of the 
distances shows a clear clustering of similar artworks, in terms of shape, 
colour, and content.

From these visualisations we can observe that the multi-scale representation is 
able to express the similarities between artworks in terms of both fine and 
coarse characteristics. Moreover, multi-scale representation makes it possible 
to express the similarity between artworks which are only similar on some 
scales (i.e., if only the fine, or only the coarse characteristics are 
similar), as shown in Figure~\ref{fig:tsnefull}.

\section{Discussion}
\label{sec:discussion}
In this work we explored the effect of incorporating scale-variance, as put 
forward by Gluckman~\cite{Gluckman2006}, in CNN and how it can be used to learn 
deep image representations that deal well with variations in image resolution, 
object size, and image scale.  The main idea behind scale-variance is that 
decomposing an image in scale-invariant components results in an incomplete 
representation of the image, as a part of the image structure is not 
scale-invariant. As stated in Section~\ref{sec:intro} Gluckman showed that 
image classification performance can be improved by using the scale variant 
image structure.  This means that a good multi-scale image representations is 
capable of capturing both the task-relevant scale-variant and scale-invariant 
image structure. To this end we presented an approach for learning 
scale-variant and scale-invariant representations by means of an ensemble of 
scale-specific CNN.  By allowing each scale-specific CNN to learn the features 
which are relevant for the task at that scale, regardless whether they are 
scale-invariant or not, we are able to construct a multi-scale representation 
that captures both scale-variant and scale-invariant image features. 

We demonstrated the effectiveness of our multi-scale CNN approach on an artist 
attribution task, on which it outperformed a single-scale CNN and was superior 
to the state-of-the-art performance on the attribution task.  Furthermore, we 
show that the best performance is achieved by combining all scales, exploiting 
the fact that scale-specific attribution performance varies greatly for 
different artists. 

Is a multi-scale approach really necessary?
Our approach requires multiple scale-specific CNNs, which may be combined into 
a single more sophisticated CNN which acquires coarse- to fine-grained 
features, using high resolution images.  However, such a network would have to 
be significantly deeper and more complex than the network used in this paper.  
Which would increase the computational cost for training and the amount of 
training data that is needed beyond what is practically feasible at this time.  
Therefore, we cannot rule out that a single sophisticated CNN may obtain a 
similar performance as our multi-scale CNN. Moreover, we suspect that such a 
network will struggle with coarse characteristics which are very dissimilar 
when observed at a fine scale, but very similar on a coarse scale, as the 
coarse scale analysis is conditioned on the fine scale analysis.  Therefore, we 
expect that a single very complex CNN will not work as well as our multi-scale 
CNN.

Additionally, we compared our approach to Multi-scale training \cite{Wu2015} 
and showed that construction a Gaussian Pyramid of the input increases 
performance and decreases the correlations between scales. While constructing 
the Gaussian Pyramid increases the computational load slightly, we believe that 
the reduced correlations between scales implies that our approach is better at 
capturing the scale variant characteristics, and is subsequently able to 
leverage these for increased performance.

Compared to previously proposed CNN architectures that deal with 
scale-variation, our approach requires many more model parameters, as the 
parameters are not shared between the single-scale CNN.  However, we consider 
this a key attribute of the approach as it enables the model to learn 
scale-variant features, and moreover, because the parameters are not shared the 
models can be trained independently and in parallel. Despite this, a potential 
downside of our approach is that we do not explicitly learn scale-invariant 
features, while they might implicitly emerge from the training procedure, 
future work on how to explicitly learn scale-variant and scale-invariant 
features is needed.

We expect that the use of multi-scale CNNs will improve performances on image 
recognition tasks that involve images with both fine and coarse-grained 
task-relevant details. Examples of such tasks are scene classification, aerial 
image analysis, and biomedical image analysis. 

\section{Conclusion}
\label{sec:conclusion}
There is a vast amount visual information to be gleaned from multi-scale images 
in which both the coarse and the fine grained details are represented.  
However, capturing all of this visual information in a deep image 
representation is non trivial. In this paper we proposed an approach for 
learning scale-variant and scale-invariant representations from high-resolution 
images. By means of a multi-scale CNN architecture consisting of multiple 
single-scale CNN, we exploit the strength of CNN in learning scale-variant 
representations, and combine these over multiple scales to encourage 
scale-invariance and improve performance. We demonstrate this by analysing the 
large amount of available details in multi-scale images for a computational 
artist attribution task, improving on the current state-of-the-art.

Moreover, we found that the representations at the various scales differ both 
in performance and in image structure learnt, and that they are complementary: 
averaging the class posteriors across all scales leads to optimal performance.  
We conclude by stating that encouraging the combined development of 
scale-invariant and scale-variant representations in CNNs is beneficial to 
image recognition performance for tasks involving image structure at varying 
scales and resolutions and merits further exploration.

\section*{Acknowledgment}
We would like to thank the anonymous reviewers for their insightful and 
constructive comments.
The research reported in this paper is performed as part of the REVIGO project, 
supported by the Netherlands Organisation for scientific research (NWO; grant 
323.54.004) in the context of the Science4Arts research program. 

\newpage
\appendix
\section{Additional comparative results}
\label{app:subresults}
\setcounter{table}{0} \begin{table}[!hp]
  \caption{Mean Class Accuracies for all possible scale combinations using the 
    Mean-scale training procedure described in \cite{Wu2015}, a `+' indicates 
  inclusion of the scale.  In bold are the combinations which lead to the best 
combined performance in each block.  The best overall score is underlined.}
\label{tab:wu_subresults}
\centering
\begin{tabular}{cccc|ccc}
\hline
256 & 512 & 1024 & 2048 & MCA & Mean recall & F-Score\\
\hline
+ & {} & {} & {} & $70.56$ & $65.74$ & $68.07$ \\
{} & + & {} & {} & $\mathbf{73.5}$ & $\mathbf{68.36}$ & $\mathbf{70.84}$ \\
{} & {} & + & {} & $65.63$ & $57.96$ & $61.56$ \\
{} & {} & {} & + & $62.03$ & $38.54$ & $47.55$ \\
\hline
+ & + & {} & {} & $75.93$ & $\mathbf{71.02}$ & $\mathbf{73.4}$ \\
+ & {} & + & {} & $75.13$ & $70.2$ & $72.58$ \\
+ & {} & {} & + & $75.68$ & $68.08$ & $71.68$ \\
{} & + & + & {} & $74.8$ & $68.51$ & $71.51$ \\
{} & + & {} & + & $\mathbf{77.8}$ & $66.2$ & $71.53$ \\
{} & {} & + & + & $68.7$ & $54.69$ & $60.9$ \\
\hline
+ & + & + & {} & $78.21$ & $\mathbf{72.94}$ & $\mathbf{75.48}$ \\
+ & + & {} & + & $\mathbf{79.16}$ & $71.95$ & $75.38$ \\
+ & {} & + & + & $77.72$ & $70.54$ & $73.95$ \\
{} & + & + & + & $77.04$ & $66.65$ & $71.47$ \\
\hline
+ & + & + & + & $\mathbf{\underline{79.98}}$ & $\mathbf{\underline{73.02}}$ & 
$\mathbf{\underline{76.34}}$ \\
\hline
\end{tabular}
\end{table}

\newpage
\bibliographystyle{IEEEtran}
\bibliography{library}

\begin{thebibliography}{10}
\expandafter\ifx\csname url\endcsname\relax
  \def\url#1{\texttt{#1}}\fi
\expandafter\ifx\csname urlprefix\endcsname\relax\def\urlprefix{URL }\fi
\expandafter\ifx\csname href\endcsname\relax
  \def\href#1#2{#2} \def\path#1{#1}\fi

\bibitem{Krizhevsky2012}
A.~Krizhevsky, I.~Sutskever, G.~Hinton,
  \href{https://papers.nips.cc/paper/4824-imagenet-classification-with-deep-convolutional-neural-networks.pdf}{{ImageNet
  Classification with Deep Convolutional Neural Networks.}}, in: Advances in
  Neural Information Processing Systems 25, 2012, pp. 1097--1105.

\bibitem{He2015}
K.~He, X.~Zhang, S.~Ren, J.~Sun, \href{http://arxiv.org/abs/1512.03385}{{Deep
  Residual Learning for Image Recognition}}, ArXiv\href
  {http://arxiv.org/abs/1512.03385} {\path{ arXiv:1512.03385}}.

\bibitem{Gong2014}
Y.~Gong, L.~Wang, R.~Guo, S.~Lazebnik,
  \href{http://arxiv.org/abs/1403.1840}{{Multi-scale Orderless Pooling of Deep
  Convolutional Activation Features}}, ArXiv (2014) 1--17\href
  {http://arxiv.org/abs/1403.1840} {\path{ arXiv:1403.1840}}.

\bibitem{LeCun1995a}
Y.~LeCun, Y.~Bengio,
  \href{http://citeseerx.ist.psu.edu/viewdoc/download?doi=10.1.1.32.9297{\&}rep=rep1{\&}type=pdf}{{Convolutional
  networks for images, speech, and time series}}, The handbook of brain theory
  and neural networks 3361 (1995) 255--258.
\newblock \href {http://dx.doi.org/10.1109/IJCNN.2004.1381049} {\path{
  doi:10.1109/IJCNN.2004.1381049}}.

\bibitem{Kheradpisheh2015}
S.~R. Kheradpisheh, M.~Ghodrati, M.~Ganjtabesh, T.~Masquelier,
  \href{http://arxiv.org/abs/1508.03929}{{Deep Networks Resemble Human
  Feed-forward Vision in Invariant Object Recognition}}, ArXiv\href
  {http://arxiv.org/abs/1508.03929} {\path{ arXiv:1508.03929}}.

\bibitem{Xu2014}
Y.~Xu, T.~Xiao, J.~Zhang, K.~Yang, Z.~Zhang,
  \href{http://arxiv.org/abs/1411.6369}{{Scale-Invariant Convolutional Neural
  Networks}}, ArXiv\href {http://arxiv.org/abs/1411.6369} {\path{
  arXiv:1411.6369}}.

\bibitem{Gonzalez:2006:DIP:1076432}
R.~C. Gonzalez, R.~E. Woods, {Digital Image Processing (3rd Edition)},
  Prentice-Hall, Inc., Upper Saddle River, NJ, USA, 2006.

\bibitem{Lindeberg1994}
T.~Lindeberg,
  \href{http://www.tandfonline.com/doi/abs/10.1080/757582976}{{Scale-space
  theory: a basic tool for analyzing structures at different scales}}, Journal
  of Applied Statistics 21~(1) (1994) 225--270.
\newblock \href {http://dx.doi.org/10.1080/757582976} {\path{
  doi:10.1080/757582976}}.

\bibitem{Gluckman2006}
J.~Gluckman,
  \href{http://ieeexplore.ieee.org/xpls/abs{\_}all.jsp?arnumber=1640869}{{Scale
  variant image pyramids}}, in: Computer Vision and Pattern Recognition, 2006,
  2006.
\newblock \href {http://dx.doi.org/10.1109/CVPR.2006.265} {\path{
  doi:10.1109/CVPR.2006.265}}.

\bibitem{Park2010}
C.~{Park, Dennis and Ramanan, Deva and Fowlkes},
  \href{http://www.springerlink.com/index/10.1007/978-3-642-15561-1}{{Multiresolution
  models for object detection}}, in: ECCV 2010, 2010, pp. 1--14.

\bibitem{Lowe2004}
D.~G. Lowe, \href{http://portal.acm.org/citation.cfm?id=996342}{{Distinctive
  image features from scale invariant keypoints}}, Int'l Journal of Computer
  Vision 60~(2).

\bibitem{Lenc2015}
K.~Lenc, A.~Vedaldi, {Understanding image representations by measuring their
  equivariance and equivalence}, Computer Vision and Pattern Recognition
  (CVPR), 2015 IEEE Conference on (2015) 991--999.

\bibitem{Le2010}
Q.~Le, J.~Ngiam, Z.~Chen, D.~H. Chia, P.~Koh,
  \href{https://papers.nips.cc/paper/4136-tiled-convolutional-neural-networks.pdf}{{Tiled
  convolutional neural networks.}}, Nips (2010) 1--9.

\bibitem{Sermanet2013}
P.~Sermanet, D.~Eigen, X.~Zhang, M.~Mathieu, R.~Fergus, Y.~LeCun,
  \href{http://arxiv.org/abs/1312.6229}{{OverFeat: Integrated Recognition,
  Localization and Detection using Convolutional Networks}}\href
  {http://arxiv.org/abs/1312.6229} {\path{ arXiv:1312.6229}}.

\bibitem{Kanazawa}
A.~Kanazawa, A.~Sharma, D.~Jacobs, {Locally Scale-Invariant Convolutional
  Neural Networks}, in: NIPS, 2014, pp. 1--11.
\newblock \href {http://arxiv.org/abs/1412.5104} {\path{ arXiv:1412.5104}}.

\bibitem{jaderberg2015}
M.~Jaderberg, K.~Simonyan, A.~Zisserman, K.~Kavukcuoglu, {Spatial Transformer
  Networks}, Nips'15 (2015) 1--14\href {http://arxiv.org/abs/1506.02025}
  {\path{ arXiv:1506.02025}}.

\bibitem{Szegedy2014}
C.~Szegedy, W.~Liu, Y.~Jia, P.~Sermanet, S.~Reed, D.~Anguelov, D.~Erhan,
  V.~Vanhoucke, A.~Rabinovich, \href{http://arxiv.org/abs/1409.4842v1}{{Going
  Deeper with Convolutions}}, arXiv (2014) 1--12\href
  {http://arxiv.org/abs/1409.4842} {\path{ arXiv:1409.4842}}.

\bibitem{Simonyan2015}
K.~Simonyan, A.~Zisserman, \href{http://arxiv.org/abs/1409.1556}{{Very Deep
  Convolutional Networks for Large-Scale Image Recognition}}, ArXiv (2015)
  1--14\href {http://arxiv.org/abs/1409.1556} {\path{ arXiv:1409.1556}}.

\bibitem{Girshick2015}
R.~Girshick, \href{http://arxiv.org/abs/1504.08083}{{Fast R-CNN}}, Arxiv\href
  {http://arxiv.org/abs/1504.08083} {\path{ arXiv:1504.08083}}, \href
  {http://dx.doi.org/10.1109/ICCV.2015.169} {\path{
  doi:10.1109/ICCV.2015.169}}.

\bibitem{Wu2015}
R.~Wu, S.~Yan, Y.~Shan, Q.~Dang, G.~Sun,
  \href{http://arxiv.org/abs/1501.02876}{{Deep Image: Scaling up Image
  Recognition}}, Arxiv (2015) 12\href {http://arxiv.org/abs/1501.02876} {\path{
  arXiv:1501.02876}}.

\bibitem{Ciresan2012a}
D.~Cireşan, U.~Meier, J.~Schmidhuber, {Multi-column Deep Neural Networks for
  Image Classification}, International Conference of Pattern
  Recognition~(February) (2012) 3642--3649.

\bibitem{Garcia2004}
C.~Garcia, M.~Delakis, {Convolutional face finder: A neural architecture for
  fast and robust face detection}, IEEE Transactions on Pattern Analysis and
  Machine Intelligence 26~(11) (2004) 1408--1423.

\bibitem{Sermanet2011}
P.~Sermanet, Y.~Lecun, {Traffic sign recognition with multi-scale convolutional
  networks}, Proceedings of the International Joint Conference on Neural
  Networks~(SEPTEMBER 2011) (2011) 2809--2813.

\bibitem{Hariharan2014}
B.~Hariharan, P.~Arbel{\'{a}}ez, R.~Girshick, J.~Malik,
  \href{http://arxiv.org/abs/1411.5752}{{Hypercolumns for Object Segmentation
  and Fine-grained Localization}}, ArXiv\href {http://arxiv.org/abs/1411.5752}
  {\path{ arXiv:1411.5752}}.

\bibitem{Hou2015}
L.~Hou, D.~Samaras, T.~Kurc, Y.~Gao,
  \href{http://arxiv.org/abs/1504.07947}{{Efficient Multiple Instance
  Convolutional Neural Networks for Gigapixel Resolution Image
  Classification}}, arXiv\href {http://arxiv.org/abs/1504.07947} {\path{
  arXiv:1504.07947}}.

\bibitem{Ciresan2012}
D.~Cireşan, U.~Meier, J.~Masci, J.~Schmidhuber,
  \href{http://dx.doi.org/10.1016/j.neunet.2012.02.023}{{Multi-column deep
  neural network for traffic sign classification}}, Neural Networks 32 (2012)
  333--338.
\newblock \href {http://arxiv.org/abs/arXiv:1202.2745v1} {\path{
  arXiv:arXiv:1202.2745v1}}, \href
  {http://dx.doi.org/10.1016/j.neunet.2012.02.023} {\path{
  doi:10.1016/j.neunet.2012.02.023}}.

\bibitem{Adelson1984}
E.~H. Adelson, C.~H. Anderson, J.~Bergen, P.~Burt, J.~M. Ogden,
  \href{https://alliance.seas.upenn.edu/{~}cis581/wiki/Lectures/Pyramid.pdf}{{Pyramid
  methods in image processing}}, RCA Engineer 29~(6) (1984) 33--41.
\newblock \href {http://dx.doi.org/10.1.1.59.9419} {\path{
  doi:10.1.1.59.9419}}.

\bibitem{Noord2015}
N.~van Noord, E.~Hendriks, E.~Postma, {Towards discovery of the artist's style:
  Learning to recognise artists by their artworks}, IEEE Signal Processing
  Magazine (2015) 1--8.

\bibitem{Johnson2008}
C.~R. {Johnson, Jr.}, E.~Hendriks, I.~J. Berezhnoy, E.~Brevdo, S.~M. Hughes,
  I.~Daubechies, J.~Li, E.~Postma, J.~Z. Wang,
  \href{http://ieeexplore.ieee.org/xpls/abs{\_}all.jsp?arnumber=4545847}{{Image
  processing for artist identification}}, IEEE Signal Processing Magazine
  (Special Section - Signal Processing in Visual Cultural Heritage)~(25) (2008)
  37--48.

\bibitem{Taylor2007}
R.~Taylor, R.~Guzman, T.~Martin,
  \href{http://www.sciencedirect.com/science/article/pii/S016786550600208X}{{Authenticating
  Pollock paintings using fractal geometry}}, Pattern Recognition Letters
  28~(6) (2007) 695--702.

\bibitem{Hughes2010a}
J.~M. Hughes, D.~J. Graham, D.~N. Rockmore,
  \href{http://www.pubmedcentral.nih.gov/articlerender.fcgi?artid=2824352{\{}{\&}{\}}tool=pmcentrez{\{}{\&}{\}}rendertype=abstract}{{Quantification
  of artistic style through sparse coding analysis in the drawings of Pieter
  Bruegel the Elder.}}, Proceedings of the National Academy of Sciences of the
  United States of America 107~(4) (2010) 1279--1283.
\newblock \href {http://dx.doi.org/10.1073/pnas.0910530107} {\path{
  doi:10.1073/pnas.0910530107}}.

\bibitem{VanderMaaten2010}
L.~J.~P. van~der Maaten, E.~O. Postma,
  \href{http://proceedings.spiedigitallibrary.org/proceeding.aspx?articleid=754635}{{Texton-Based
  Analysis of Paintings}}, in: SPIE Optical Imaging and Applications, Vol.
  7798, 2010.

\bibitem{Li2012}
J.~Li, L.~Yao, E.~Hendriks, J.~Z. Wang,
  \href{http://www.ncbi.nlm.nih.gov/pubmed/22516651}{{Rhythmic brushstrokes
  distinguish van Gogh from his contemporaries: findings via automated
  brushstroke extraction.}}, IEEE transactions on pattern analysis and machine
  intelligence 34~(6) (2012) 1159--1176.
\newblock \href {http://dx.doi.org/10.1109/TPAMI.2011.203} {\path{
  doi:10.1109/TPAMI.2011.203}}.

\bibitem{Abry2013}
P.~Abry, H.~Wendt, S.~Jaffard,
  \href{http://linkinghub.elsevier.com/retrieve/pii/S0165168412000308}{{When
  Van Gogh meets Mandelbrot: Multifractal classification of painting's
  texture}}, Signal Processing 93~(3) (2013) 554--572.
\newblock \href {http://dx.doi.org/10.1016/j.sigpro.2012.01.016} {\path{
  doi:10.1016/j.sigpro.2012.01.016}}.

\bibitem{Hendriks2008}
E.~Hendriks, S.~Hughes, {Van Gogh ' s brushstrokes : marks of authenticity ?},
  Art, Conservation, and Authenticities: Material, Concept, Context (2008)
  57--62.

\bibitem{Hendriks2012}
D.~H. Johnson, E.~Hendriks, C.~R.~J. Jr,
  \href{http://www-ece.rice.edu/{~}dhj/canvas.pdf}{{Interpreting canvas weave
  matches}}, Art Matters (2012) 53--61.

\bibitem{Mensink2014}
T.~Mensink, J.~V. Gemert, \href{http://dl.acm.org/citation.cfm?id=2578791}{{The
  Rijksmuseum Challenge: Museum-Centered Visual Recognition}}, Proceedings of
  International Conference on Multimedia Retrieval (2014) 2--5.

\bibitem{Karayev2013}
S.~Karayev, M.~Trentacoste, H.~Han, A.~Agarwala, T.~Darrell, A.~Hertzmann,
  H.~Winnemoeller, \href{http://arxiv.org/abs/1311.3715}{{Recognizing Image
  Style}}, ArXiv\href {http://arxiv.org/abs/1311.3715} {\path{
  arXiv:1311.3715}}.

\bibitem{Saleh2015}
B.~Saleh, A.~Elgammal, \href{http://arxiv.org/abs/1505.00855}{{Large-scale
  Classification of Fine-Art Paintings: Learning The Right Metric on The Right
  Feature}}, arXiv (2015) 21\href {http://arxiv.org/abs/1505.00855} {\path{
  arXiv:1505.00855}}.

\bibitem{springenberg2015}
J.~T. Springenberg, A.~Dosovitskiy, T.~Brox, M.~Riedmiller, {Striving for
  Simplicity: The All Convolutional Net}, in: ICLR, 2015.
\newblock \href {http://arxiv.org/abs/1412.6806} {\path{ arXiv:1412.6806}}.

\bibitem{Zeiler2014}
M.~Zeiler, R.~Fergus,
  \href{http://link.springer.com/chapter/10.1007/978-3-319-10590-1{\_}53}{{Visualizing
  and understanding convolutional networks}}, ECCV 2014 8689 (2014) 818--833.
\newblock \href {http://dx.doi.org/10.1007/978-3-319-10590-1} {\path{
  doi:10.1007/978-3-319-10590-1}}.

\bibitem{Ogden1985}
M.~Ogden, H.~Adelson, R.~Bergen, J.~Burt, {Pyramid-based computer graphics},
  RCA Engineer.

\bibitem{Maaten2008}
L.~van~der Maaten, G.~Hinton, {Visualizing Data using t-SNE.}, Journal of
  Machine Learning Research 9 (2008) 2579--2605.

\end{thebibliography}

\end{document}